\documentclass[10pt,twocolumn,letterpaper]{article}

\usepackage{cvpr}              
\usepackage{lipsum}
\usepackage{amssymb}
\usepackage{multirow}
\usepackage[<options>]{animate}
\usepackage{graphicx}
\usepackage[dvipsnames]{xcolor}

\usepackage{pifont}
\newcommand{\modelname}{PIA}
\newcommand{\benchname}{AnimateBench}

\usepackage{lipsum}
\newcommand\blfootnote[1]{%
  \begingroup
  \renewcommand\thefootnote{}\footnote{#1}%
  \addtocounter{footnote}{-1}%
  \endgroup
}

\def\paperTitle{
PIA: Your Personalized Image Animator \\
via Plug-and-Play Modules in Text-to-Image Models  }

\definecolor{cvprblue}{rgb}{0.21,0.49,0.74}
\usepackage[pagebackref,breaklinks,colorlinks,citecolor=cvprblue]{hyperref}
\usepackage[accsupp]{axessibility} 
\usepackage[hypcap=false]{caption}

\def\paperID{6821} 
\def\confName{CVPR}
\def\confYear{2024}

\title{\paperTitle}

\author{Yiming Zhang\textsuperscript{\rm 1,2,*}\quad
Zhening Xing\textsuperscript{\rm 1,*}\quad
Yanhong Zeng\textsuperscript{\rm 1,$\dagger$}\quad
Youqing Fang\textsuperscript{\rm 1}\quad
Kai Chen\textsuperscript{\rm 1,$\dagger$}
\\
\vspace{-0.6em}\\
$^1$Shanghai Artificial Intelligence Laboratory \quad
$^2$Dalian University of Technology
\\
\vspace{-0.6em}\\
\href{https://github.com/open-mmlab/PIA}{https://github.com/open-mmlab/PIA}
}

\begin{document}

\newcommand{\figoverview}{
    \begin{figure*}[t]
        \includegraphics[width=\linewidth]{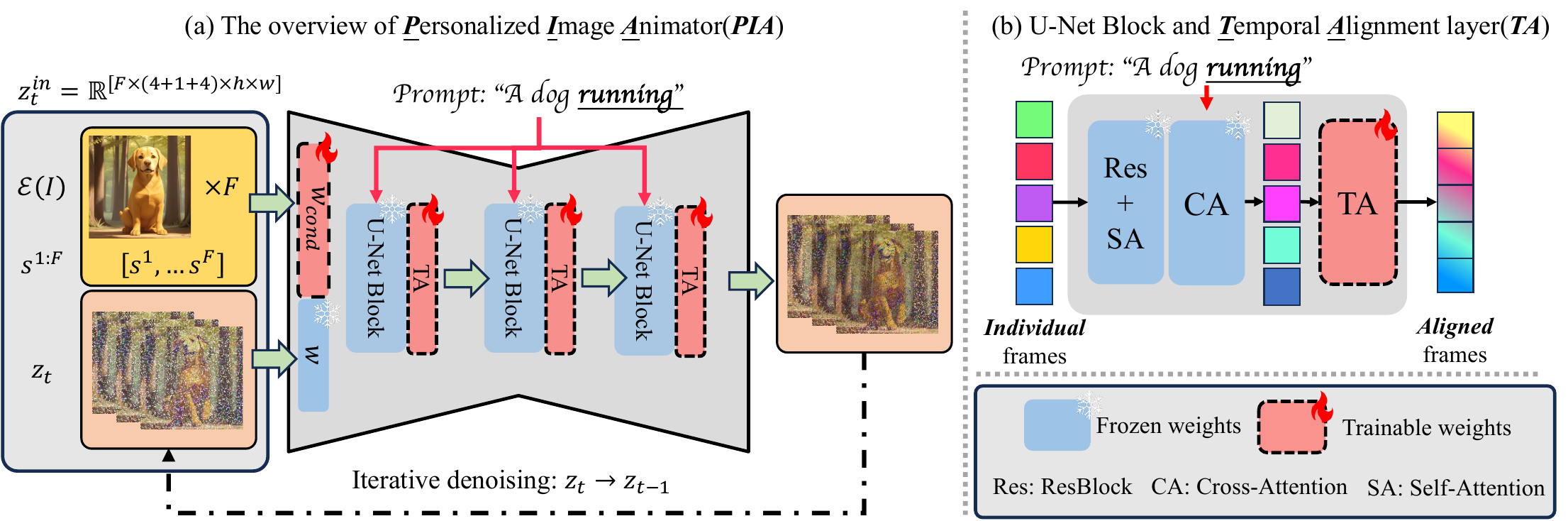}
        \caption{
            \textbf{Personalized Image Animator (\modelname)}.
            As shown in (a), \modelname\ consists of a text-to-image (T2I) model, well-trained temporal alignment layers (TA), and a new condition module $\mathcal{W}_{cond}$ responsible for encoding the condition image $z^I$ and inter-frame affinity $s^{1:F}$. In particular, the T2I model consists of U-Net blocks, including a ResBlock (Res) \cite{he2016resnet}, a self-attention layer (SA), and a cross-attention layer (CA), as depicted in (b).
            During training, the condition module learns to leverage the affinity hints and incorporate appearance information from the condition images, facilitating image alignment and enabling a stronger emphasis on motion-related alignment. 
        }
        \label{fig:overview}
    \end{figure*}
}

\newcommand{\figConInExp}{
    \begin{figure}[t]
        \includegraphics[width=1\columnwidth]{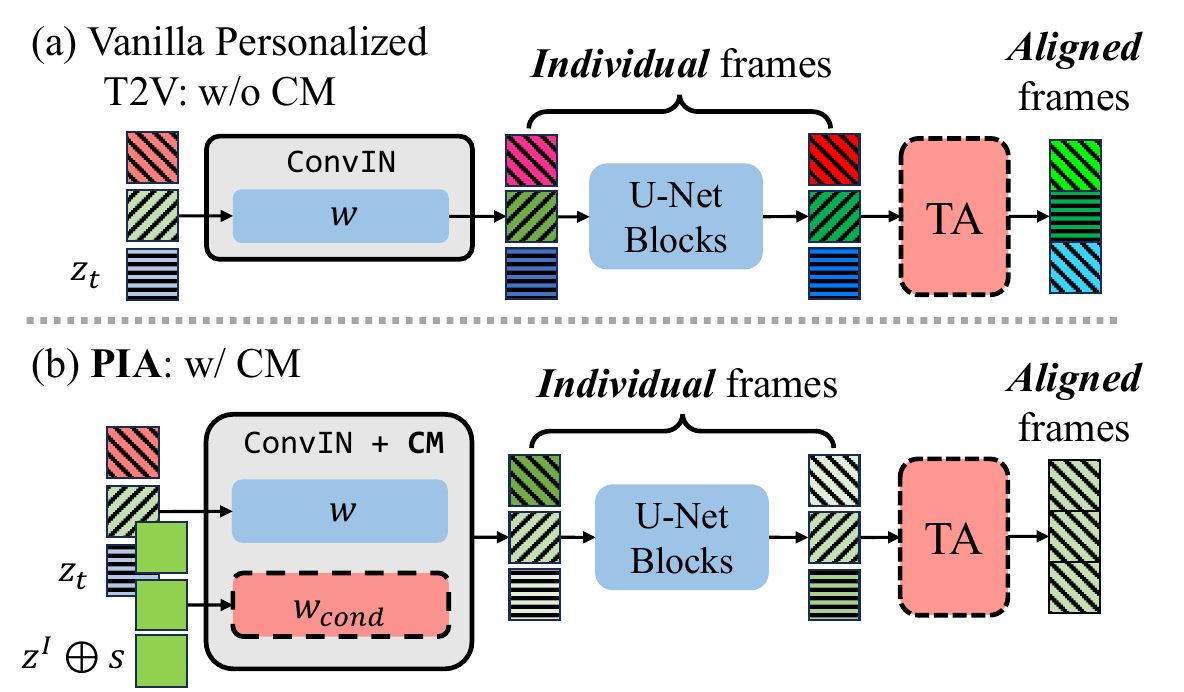}
        \caption{
            \textbf{Illustration of the condition module (CM).}
            A vanilla personalized T2V model (shown in (a)) needs to align both the appearance and motion of individual frames simultaneously.
            PIA with CM (shown in (b)) can borrow appearance information from condition image $z^I$ with affinity hints $s$, easing the challenge of both appearance and motion alignment.
            We use the color and strip to denote appearance and motion, respectively.
        }
        \vspace{-3mm}
        \label{fig:explain_cond}
    \end{figure}
}

\newcommand{\figanimatebench}{
\begin{figure}[t]
\includegraphics[width=\linewidth]{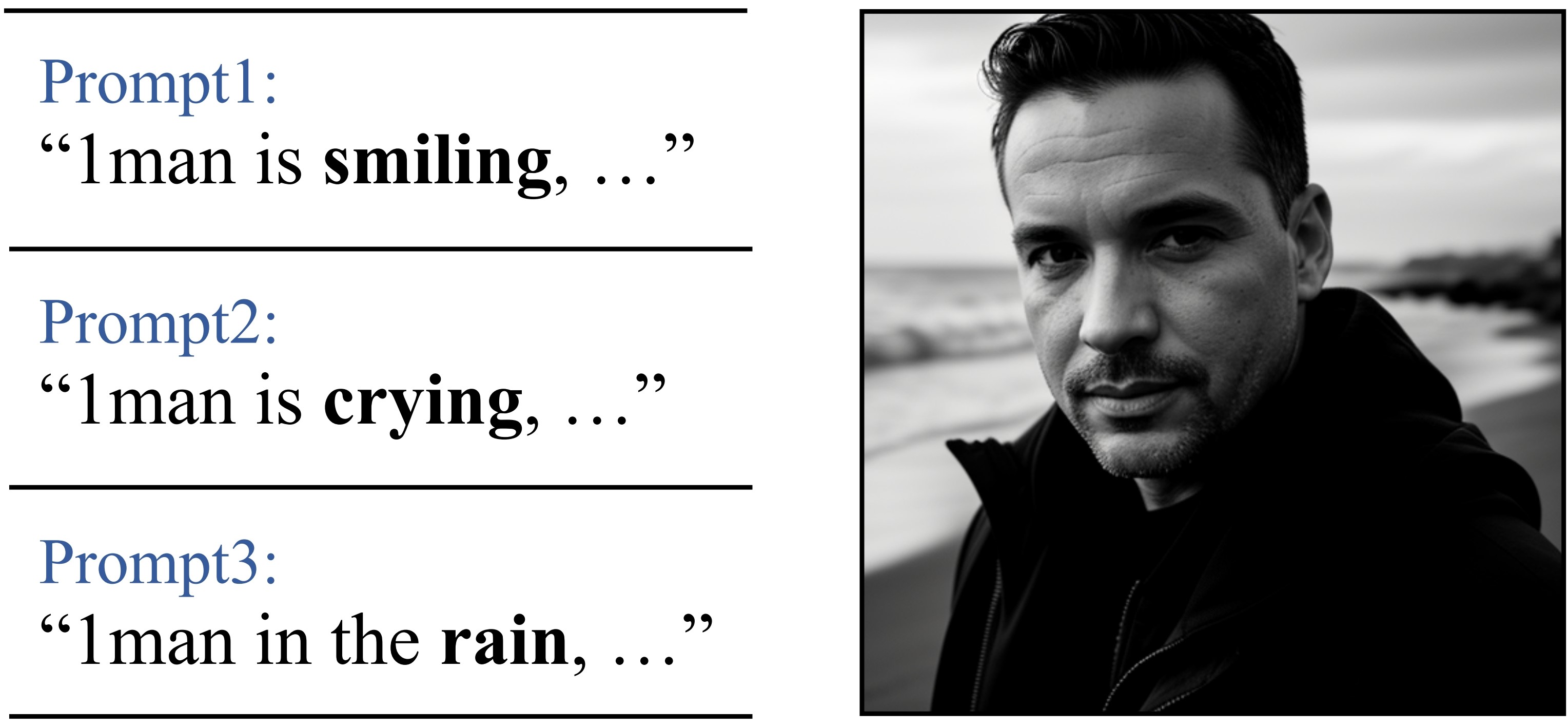}
\caption{An example of \textbf{AnimateBench}. The images in \benchname\ are carefully crafted using a set of collected personalized text-to-image models. Each image has three carefully designed prompts, describing the following motions that the image probably happens within a single short shot.}
\label{fig:animatebench}
\end{figure}
}

\newcommand{\figvisual}{
\begin{figure*}[t]
\includegraphics[width=\linewidth]{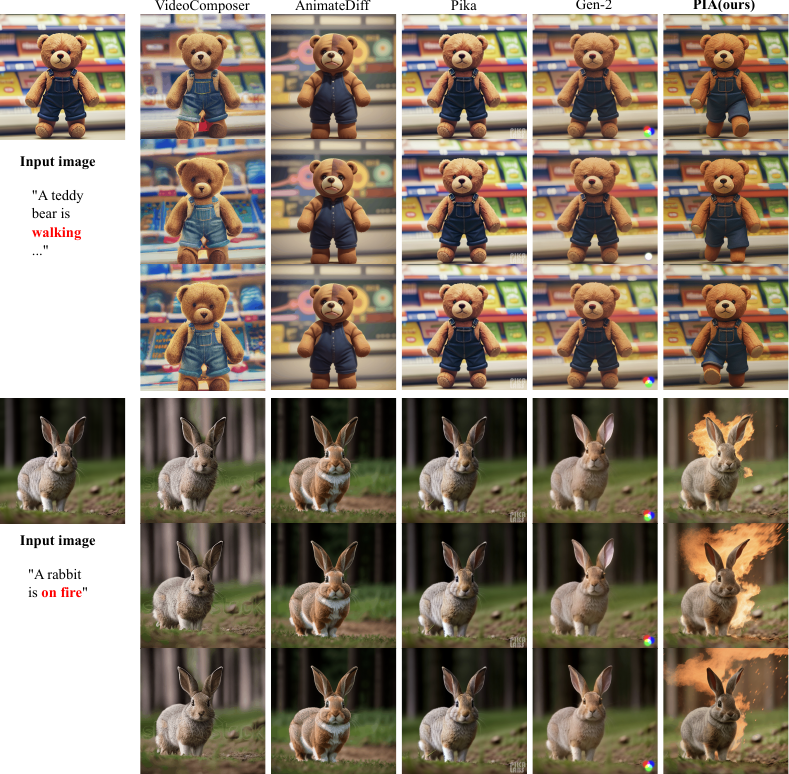}
\caption{\textbf{Qualitative comparison with state-of-the art approaches.} 
Compared with other methods, PIA shows excellent motion controllability and strong image alignment. Specifically, in the first case, PIA generates a ``walking" motion for the toy bear (in its feet), while other methods can only remain static frames, showing a lack of motion controllability. In the second case, PIA adds a new element, \ie, fire, with realistic motion. We show more video cases in supplementary materials due to the file size limit of the main paper.}
\label{fig:visual}
\end{figure*}
}

\newcommand{\figuserstudy}{
\begin{figure}[t]
\includegraphics[width=\linewidth]{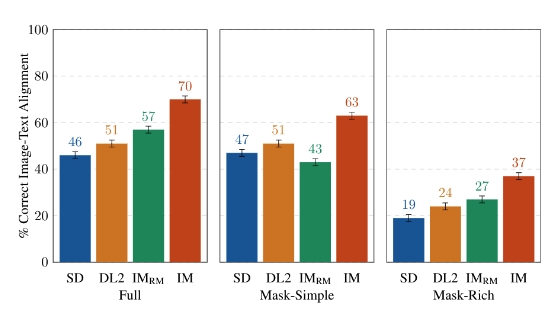}
\caption{User study}
\label{fig:userstudy}
\end{figure}
}

\newcommand{\figlimit}{
\begin{figure}[t]
\includegraphics[width=\linewidth]{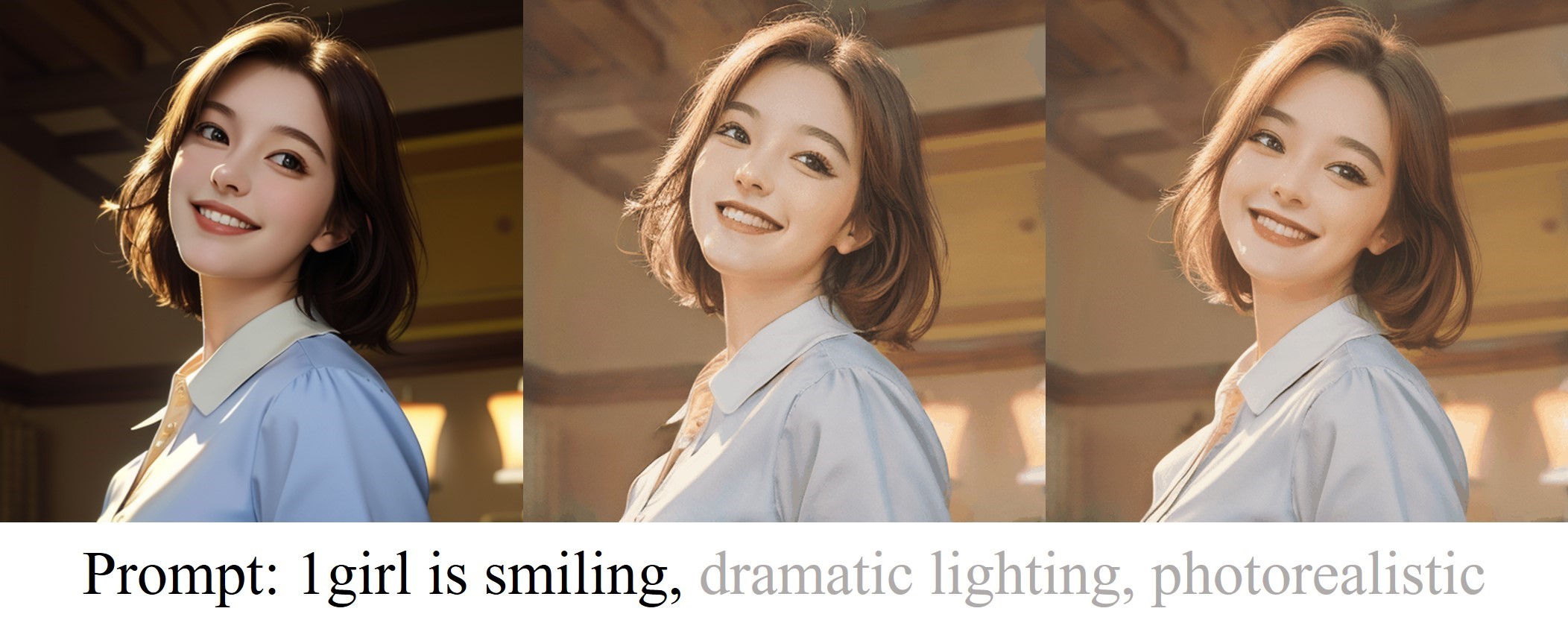}
\vspace{-5mm}
\caption{\textbf{Limitation.} When the input image is far from the training dataset domain, PIA is prone to generate videos with significant shifts of color.}
\label{fig:limit}
\end{figure}
}

\newcommand{\figmotion}{
\begin{figure}[t]
\includegraphics[width=\linewidth]{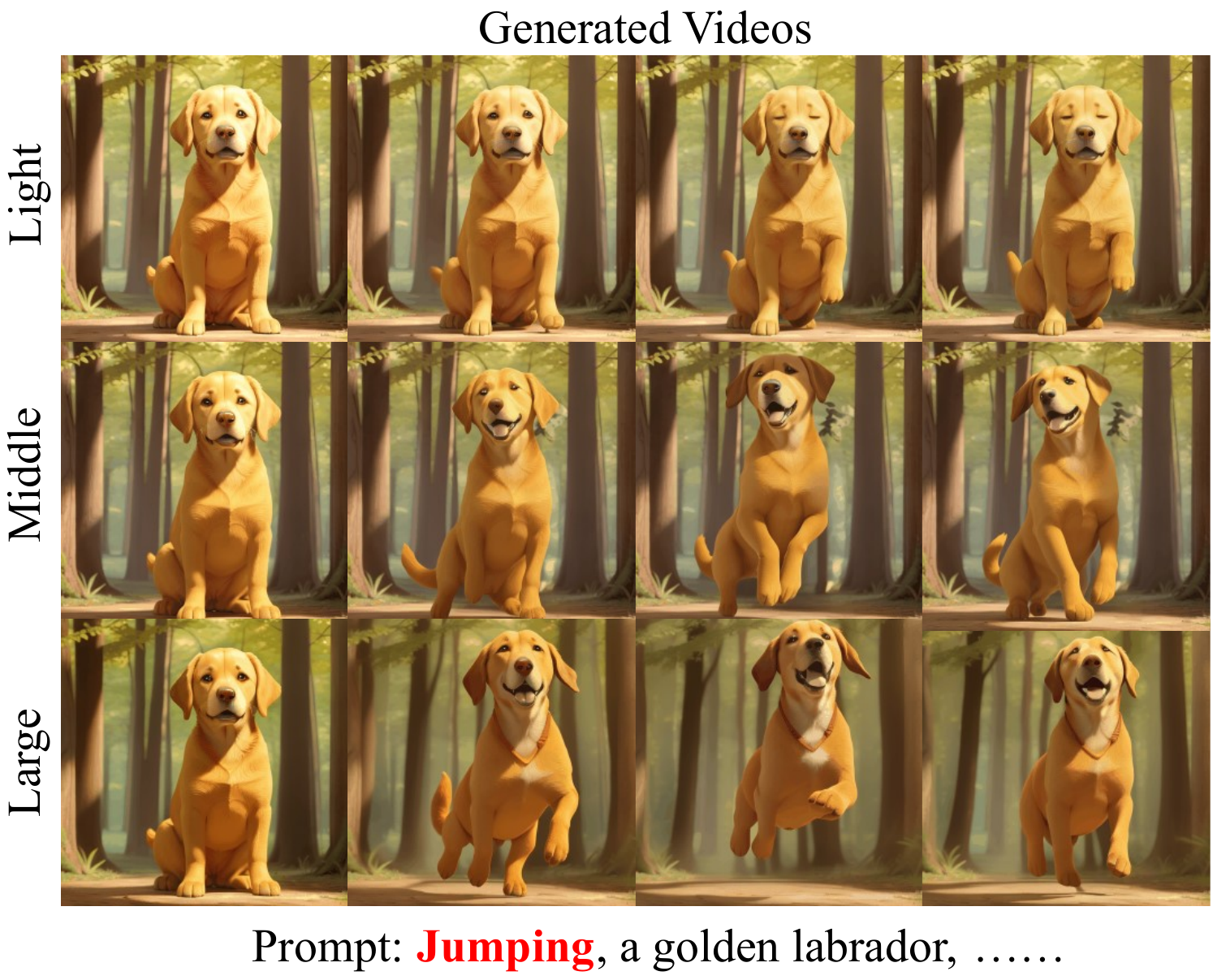}
\vspace{-5mm}
\caption{\textbf{Motion Magnitude Controllability.} 
PIA enables users to adjust the motion magnitude as light, middle, or large by setting the input affinity information as different values.}
\vspace{-3mm}
\label{fig:motion}
\end{figure}
}

\newcommand{\figpos}{
\begin{figure}[t]
\includegraphics[width=\linewidth]{figs/ours/position.jpg}
\caption{Input image on the different position in video clips}
\label{fig:position}
\end{figure}
}

\newcommand{\figprompt}{
\begin{figure}[t]
\includegraphics[width=\linewidth]{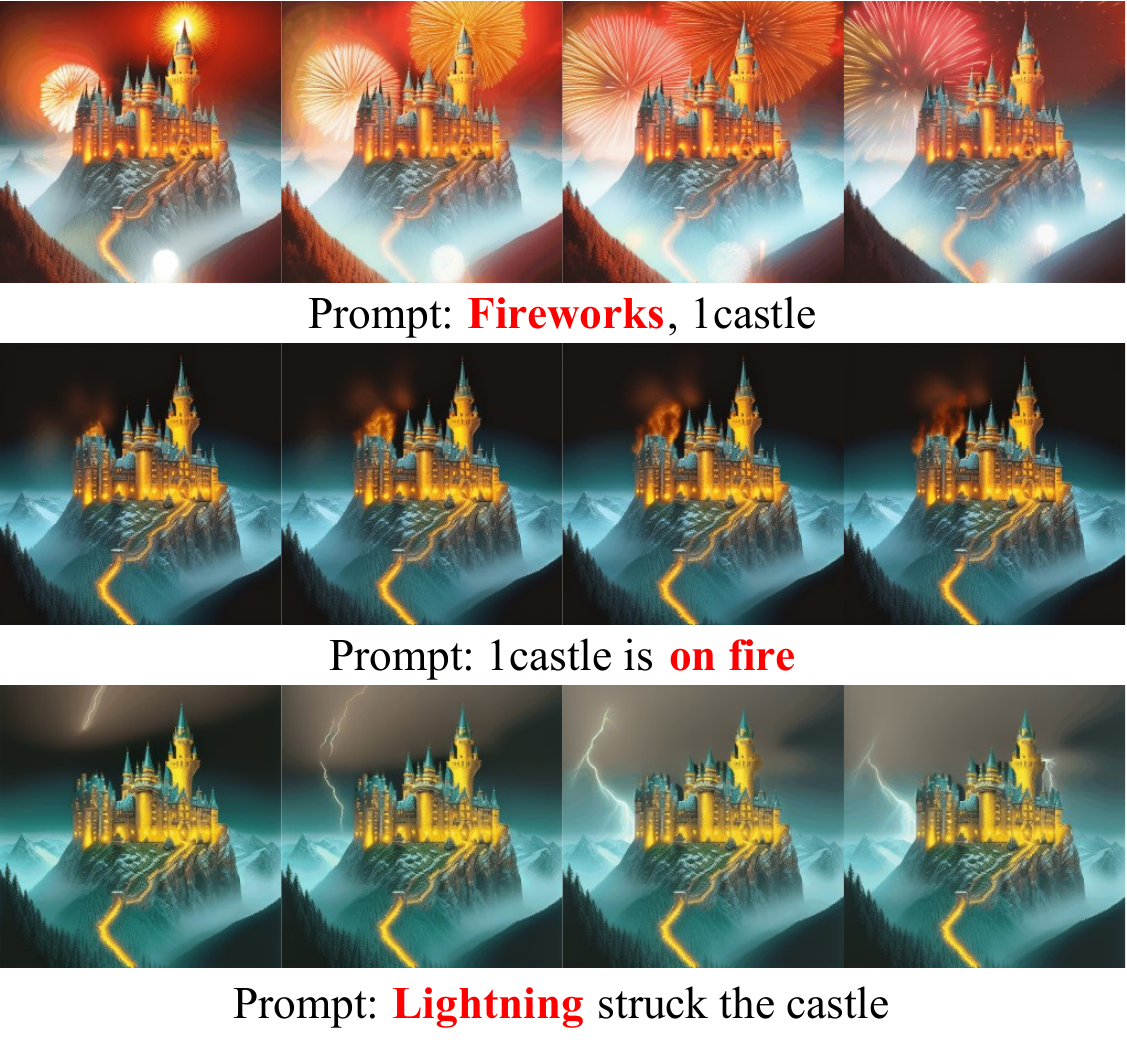}
\vspace{-7mm}
\caption{\textbf{Motion Control by Text Prompt.} \modelname\ can effectively capture the motion-related guidance in text prompt and add realistic related motion in the results.}
\label{fig:prompt}
\end{figure}
}

\newcommand{\figstyle}{
\begin{figure}[t]
\includegraphics[width=\linewidth]{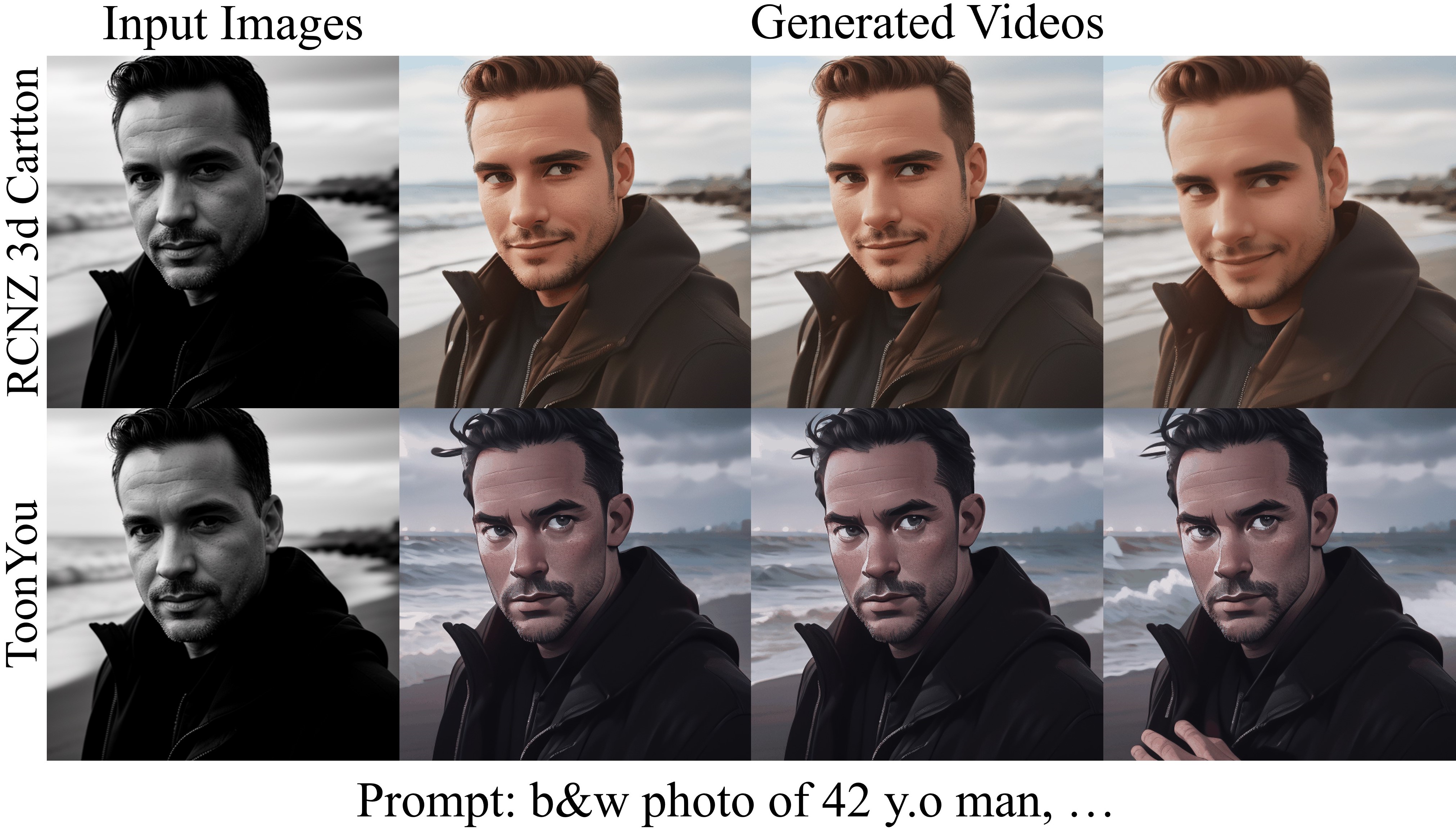}
\vspace{-7mm}
\caption{\textbf{Style transfer.}  Given the same input image and text prompt, PIA can generate different styles of videos like style transfer by using different personalized text-to-image models (\eg, RCNZ 3d Cartoon and ToonYou in this case).}
\label{fig:style}
\end{figure}
}

\newcommand{\figabtable}{
\begin{figure}[t]
\includegraphics[width=\linewidth]{figs/animatebench_table.jpg}
\vspace{-7mm}
\caption{\textbf{Model List.}  
\benchname\ consists of base models and LoRAs which contains variety of concept, style and content.}
\label{fig:abtable}
\end{figure}
}

\newcommand{\figattention}{
\begin{figure}[t]
\includegraphics[width=\linewidth]{figs/limitation.jpg}
\vspace{-5mm}
\caption{visualization of attention maps and features}
\label{fig:attention}
\end{figure}
}

\newcommand{\figsuppluserstudy}{
\begin{figure*}[t]
\centering
\includegraphics[width=0.7\linewidth]{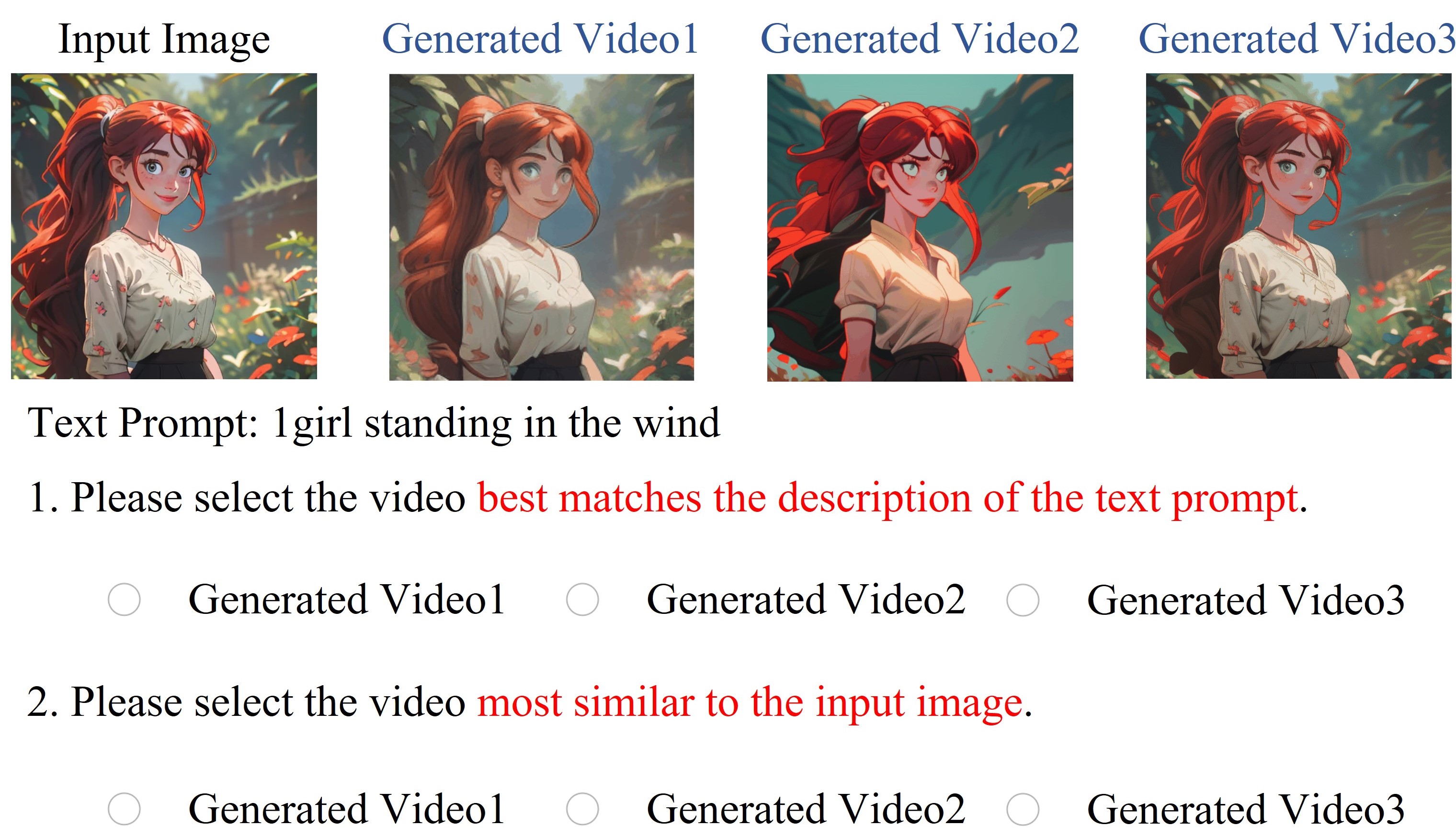}
\caption{\textbf{User Study.} Example of user study questionnaires.}
\label{fig:suppluserstudy}
\end{figure*}
}

\newcommand{\tabquant}{
\begin{table}[t]
\begin{center}
\resizebox{0.5\textwidth}{!}{
    \begin{tabular}{ccccccc}
    \toprule[1.5pt]
    \multirow{2}{*}{Methods} & \multicolumn{2}{c}{AnimateBench} & \multicolumn{2}{c}{WebVid}  & \multicolumn{2}{c}{User Study} \\
     ~  & image  & text   & image  & text & image & text    \\\hline
    VideoComposer \cite{wang2023videocomposer} & 90.10 & 66.17  & 85.03  & 75.52   & 0.180 & 0.110  \\\hline
    AnimateDiff \cite{guo2023animatediff}  & 89.72 & 68.41  & 72.82 & 72.86    & 0.295  & 0.220     \\\hline
    PIA (ours) & \textbf{93.44} & \textbf{68.74}   &\textbf{85.11}  &\textbf{76.59}  & \textbf{0.525} & \textbf{0.670} \\
    \bottomrule[1.5pt]
    \end{tabular}
}
\end{center}
\vspace{-3mm}
\caption{Quantitative comparison with state-of-the-art approaches.}
\label{table:quant}
\end{table}
}

\newcommand{\figaffinity}{
\begin{figure}[t]
\includegraphics[width=\linewidth]{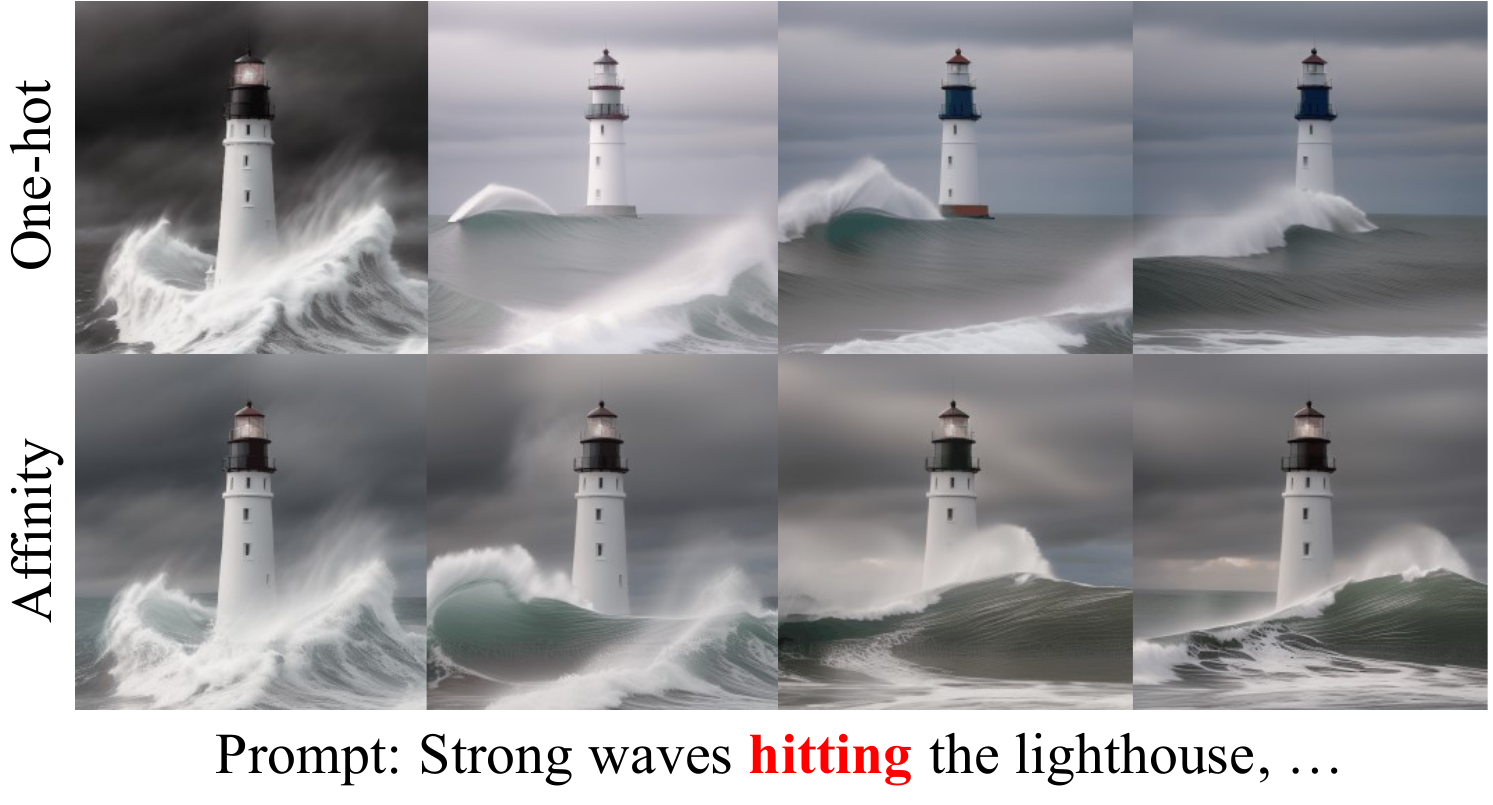}
\vspace{-5mm}
\caption{\textbf{Ablation study on the design of inter-frame affinity.} 
Compared with the one-hot condition design, the design of inter-frame affinity in PIA shows better image alignment.}
\label{fig:affinity}
\end{figure}
}

\def\tabsim{
	\begin{table}[t]
		\begin{center}
			\begin{tabular}{ccc}
    \toprule[1.5pt]
        Similarity channel& Image CLIP Score & Text CLIP Score & & \\
				\hline
		One-hot  & 21.09  & \textbf{67.87}\\
				\hline
        Distance & 21.96 & 67.68\\
        \hline
		Similarity  & \textbf{22.92}  &66.94 \\
    \bottomrule[1.5pt]
			\end{tabular}
		\end{center}
  \vspace{-5mm}
		\caption{Different construction of similarity channel.}
		\label{table:sim}
	\end{table}
}

\newcommand{\figsuppbenchcase}{
\begin{figure}[t]
\includegraphics[width=\linewidth]{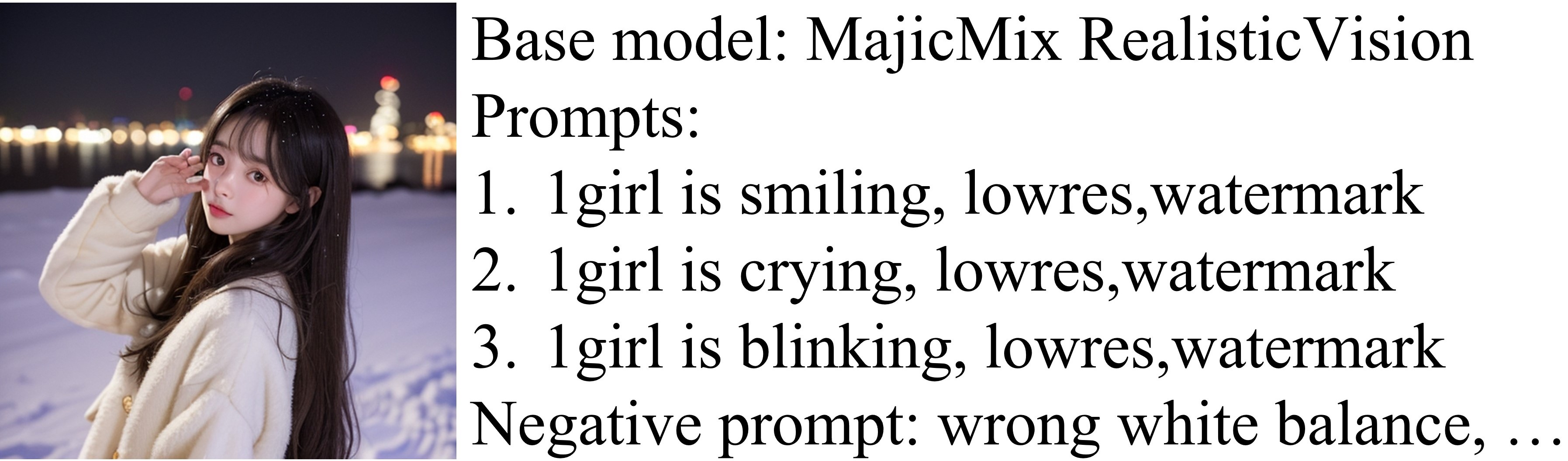}
\caption{\textbf{\benchname\ case.} Each curated personalized image corresponds to a personalized text-to-image model and three tailored motion-related text prompts.}
\label{fig:benchcase}
\end{figure}
}

\newcommand{\figonlyconditionmodule}{
\begin{figure}[t]
\includegraphics[width=\linewidth]{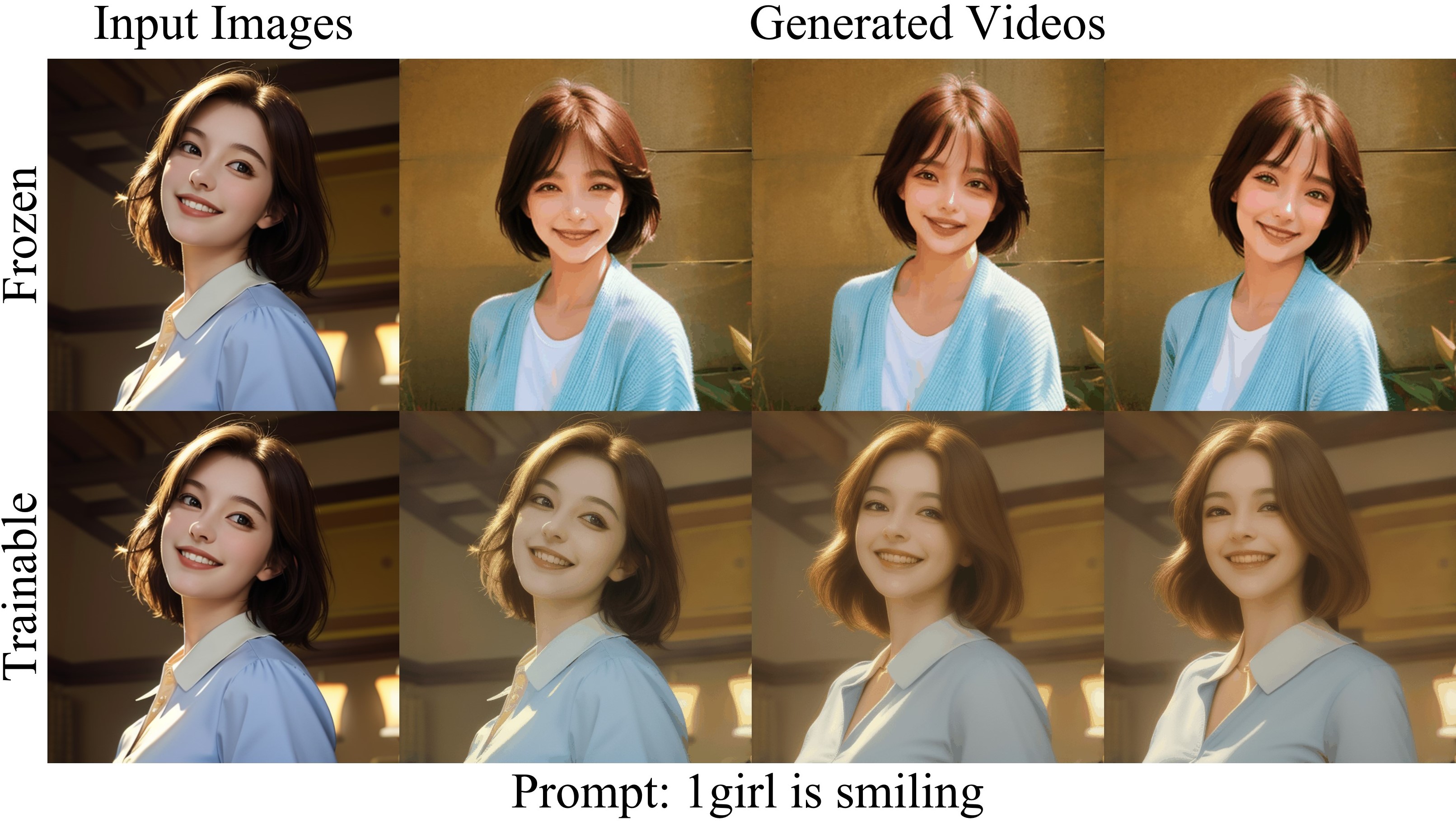}
\caption{\textbf{Ablation study for fine-tuning the Temporal Alignment Layers.} Pre-trained temporal alignment layers fail to align the condition frame in generated videos. \modelname\ fine-tunes both the condition module and the temporal alignment layers, leading to better preservation of the information in the condition frames.}
\label{fig:onlyconditionmodule}
\end{figure}
}

\newcommand{\figaffinityablation}{
\begin{figure}[t]
\includegraphics[width=\linewidth]{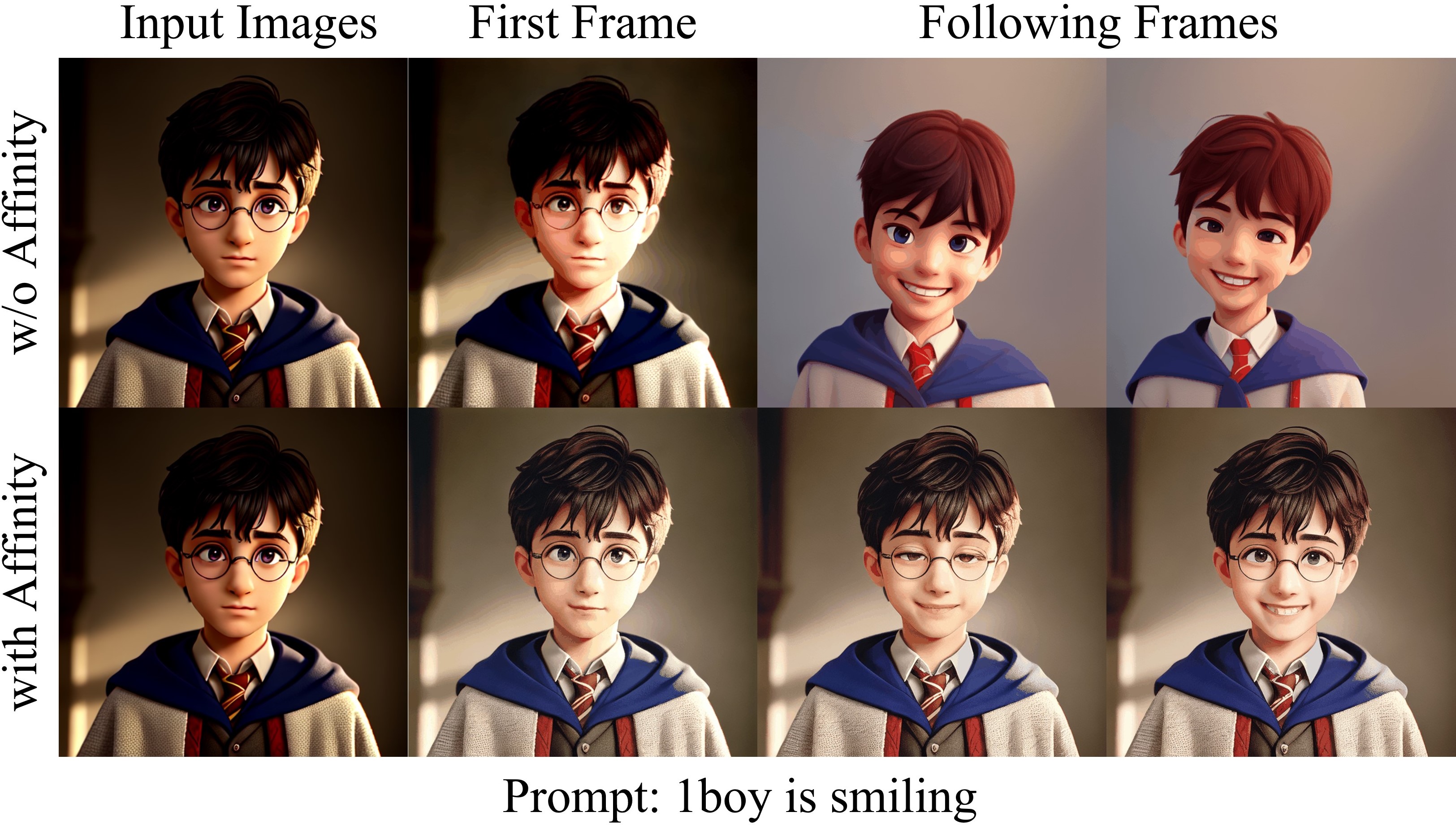}
\caption{\textbf{Ablation study for Inter-frame Affinity.} Without an affinity hint, the generated videos become incoherent and may change significantly after the first given frame. With the inter-frame affinity as inputs, \modelname\ is able to animate images that are faithful to the condition frame.}
\label{fig:affinityablation}
\end{figure}
}

\newcommand{\figfeaturemap}{
\begin{figure}[t]
\includegraphics[width=\linewidth]{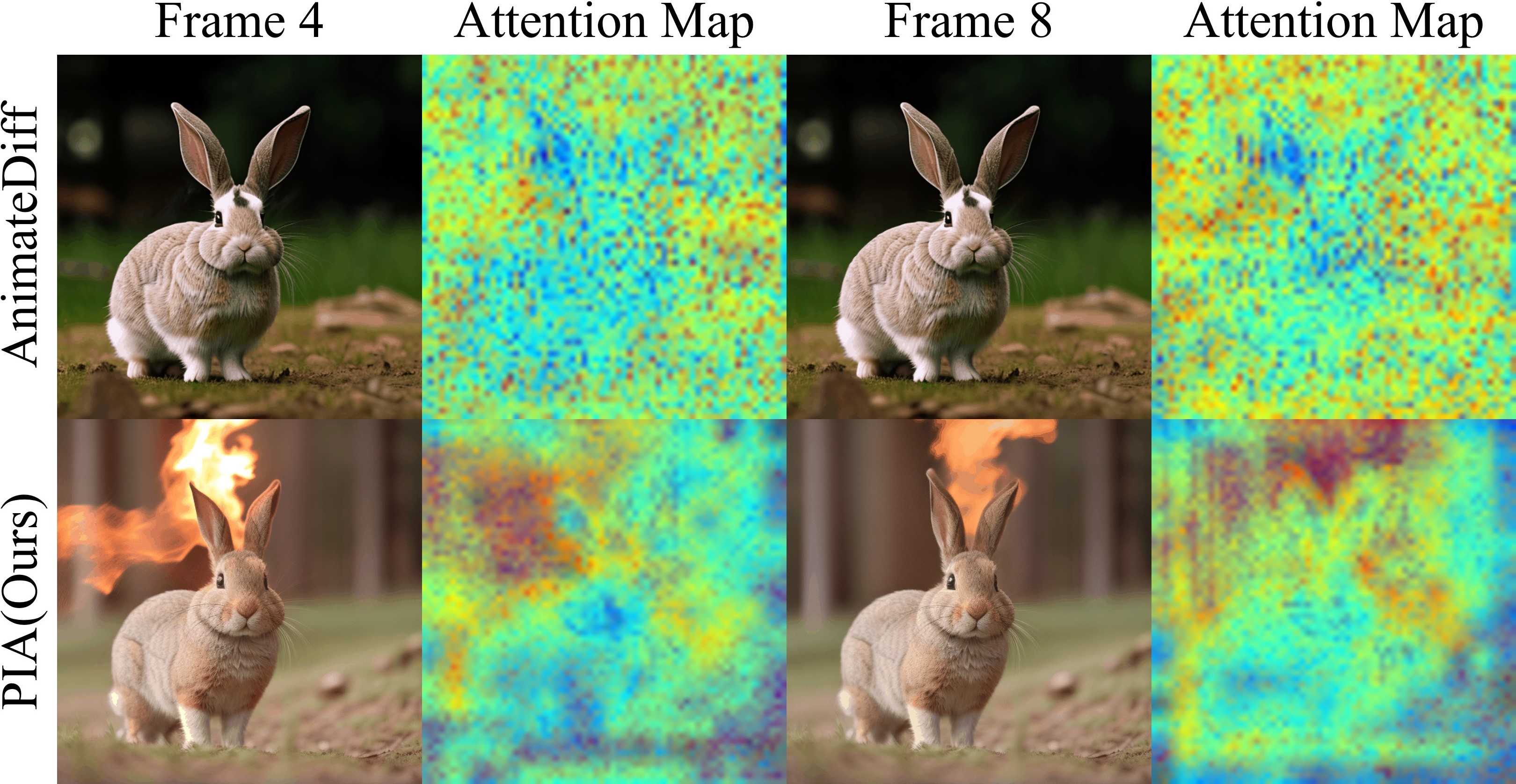}
\caption{
    \textbf{Visualization of Cross-attention map.} 
    We generate the video using prompt \textit{`a rabbit is on fire'} and visualize the cross-attention map corresponding to the token \textit{`fire'} for both AnimateDiff \cite{guo2023animatediff} and our own method.
    In PIA, token \textit{`fire'} shows more accurate attention to the shape of flames, while in AnimateDiff, the token randomly attends to the entire context.
    This demonstrates the superior motion alignment performance of our method.
}\label{fig:featuremap}
\end{figure}
}

\newcommand{\figipadapter}{
\begin{figure}[t]
\includegraphics[width=\linewidth]{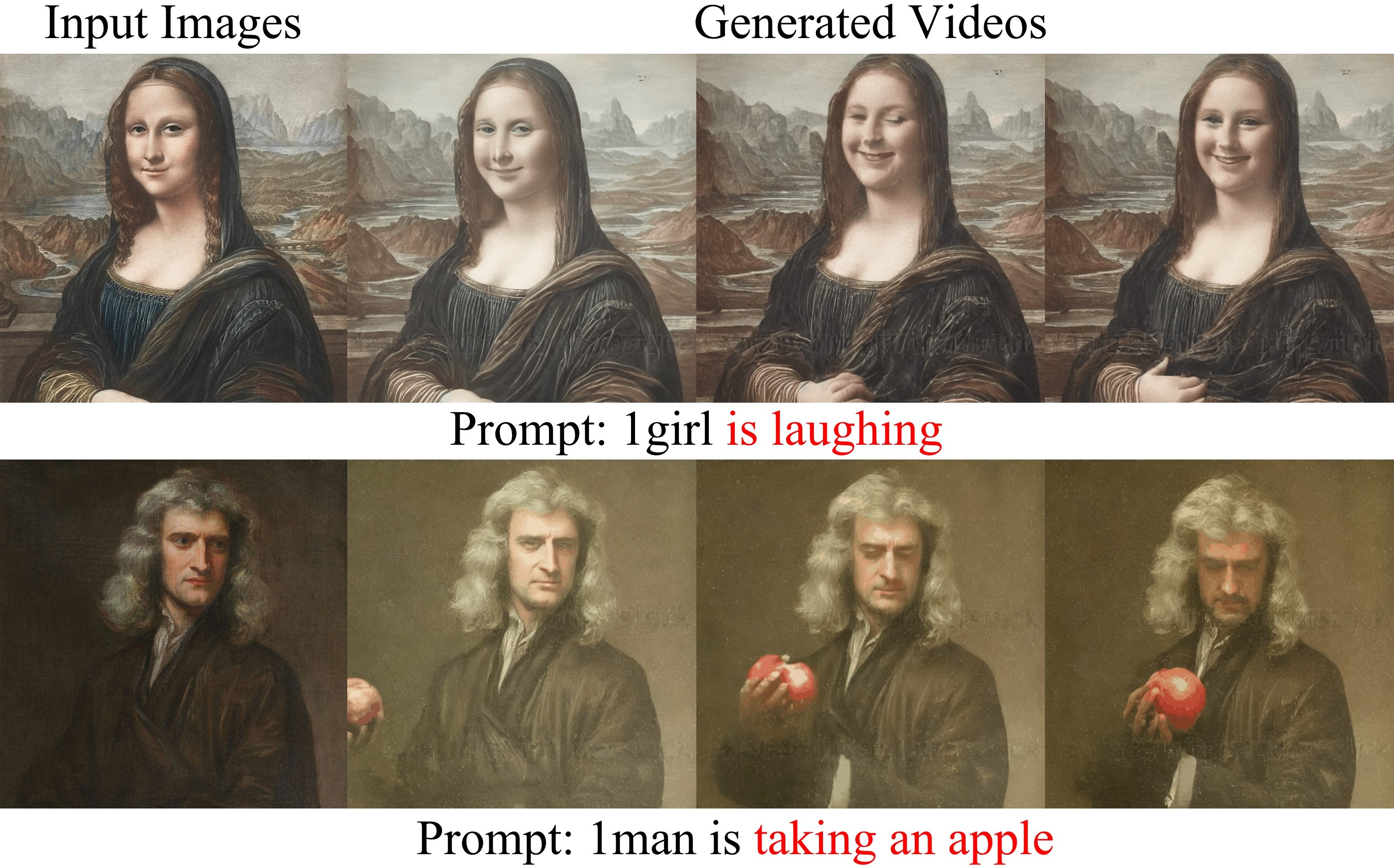}
\caption{\textbf{Using PIA to animate open-domain images.}Without
providing personalized T2I models, PIA is able to animate the
open-domain images with realistic motions by text while preserving
the details and identity in condition frame with IP-Adapter\cite{ye2023ip-adapter}}
\label{fig:ipadapter}
\end{figure}
}

\def\tabmodellist{
	\begin{table}[t]
		\begin{center}
			\begin{tabular}{lcc}
 \toprule[1.5pt]
Model Name         & Feature    & Type       \\
\hline
Realistic Vision   & Realistic  & Base Model \\
ToonYou            & 2D Cartoon & Base Model \\
RCNZ Cartoon 3d    & 3D Cartoon & Base Model \\
Lyriel             & Stylistic  & Base Model \\
majicMIX realistic & Realistic  & Base Model \\
FilmVelvia         & Realistic  & LoRA       \\
TUSUN              & Concept    & LoRA       \\   
\bottomrule[1.5pt]
\end{tabular}
		\end{center}
  \vspace{-5mm}
		\caption{\textbf{Model List.} AnimateBench consists of base models and LoRAs which contains variety of concept, style and content.}
		\label{table:modellist}
	\end{table}
}

\newcommand{\figcomplex}{
\begin{figure}[t]
\includegraphics[width=\linewidth]{figs/ours/complex.gif}
\caption{\textbf{Animation with complex prompts.} Left prompt: 1girl walking, smiling and fireworks.
Right prompt: a golden labrador jumping, smiling, on fire}
\label{fig:complex}
\end{figure}
}

\newcommand{\figcomposite}{
\begin{figure}[t]
\includegraphics[width=\linewidth]{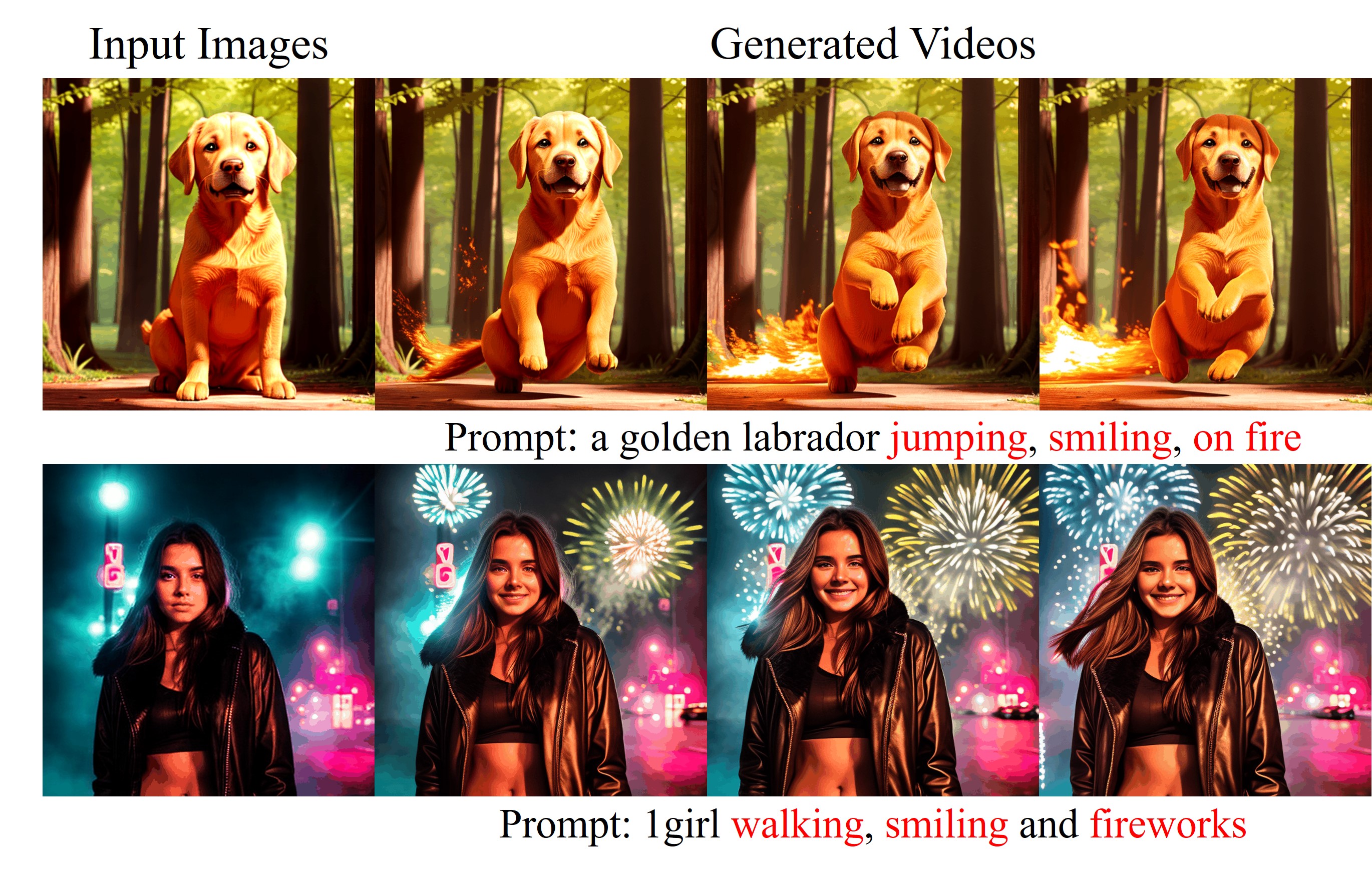}
\caption{\textbf{Animation with complex prompts.} PIA achieves improved motion controllability. Even with the complex text prompt, PIA can correspondingly generate composite animations.}
\label{fig:composite}
\end{figure}
}


\twocolumn[{%
\renewcommand\twocolumn[1][]{#1}%
\maketitle
\begin{center}
    \vspace{-6mm}
    \includegraphics[width=\linewidth]{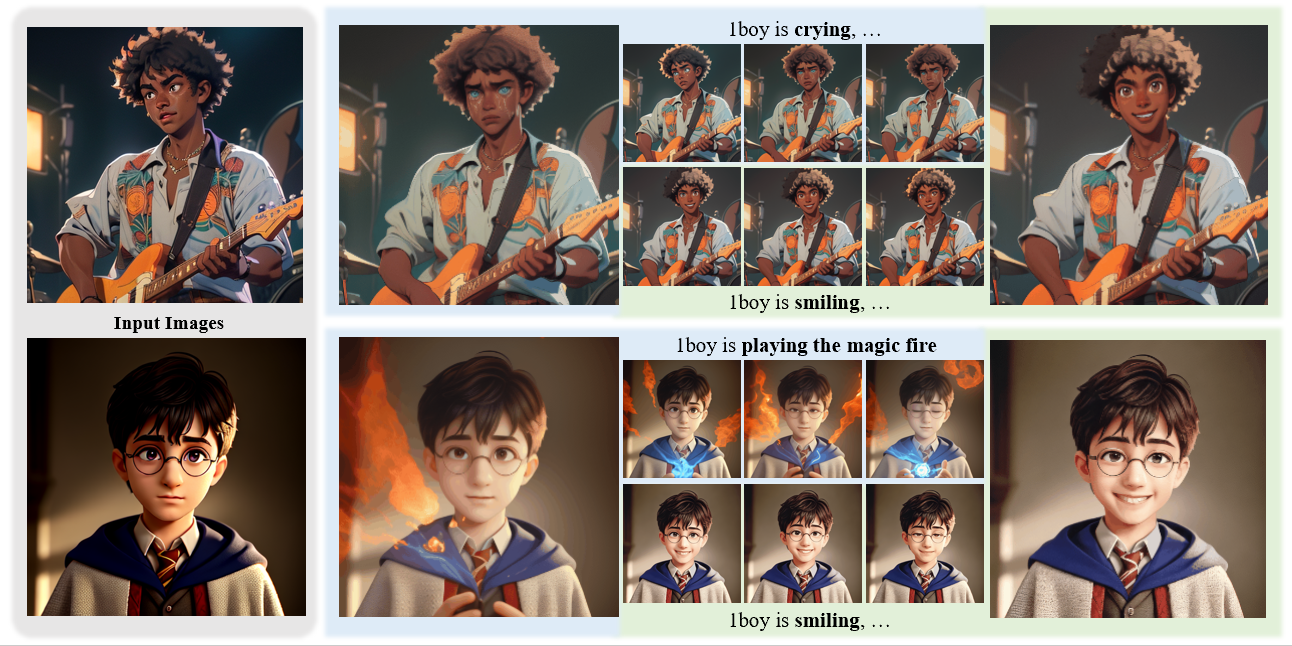}
    \vspace{-7mm}
    \captionof{figure}{Given an elaborated image generated by a personalized text-to-image model, the proposed \textbf{P}ersonalized \textbf{I}mage \textbf{A}nimator (\textbf{\modelname}) animates it with realistic motions according to different text prompts while preserving the original distinct styles and high-fidelity details. We recommend using Abode Arobat and clicking the images to play the animation clips. [Best viewed in color with zoom-in]}
\end{center}%
}]

\begin{abstract}
\blfootnote{\textsuperscript{$*$} denotes equal contribution, \textsuperscript{$\dagger$} denotes corresponding author.} 
Recent advancements in personalized text-to-image (T2I) models have revolutionized content creation, empowering non-experts to generate stunning images with unique styles. While promising, animating these personalized images with realistic motions poses significant challenges in preserving distinct styles, high-fidelity details, and achieving motion controllability by text. In this paper, we present \textbf{\modelname}, a \textbf{P}ersonalized \textbf{I}mage \textbf{A}nimator that excels in aligning with condition images, achieving motion controllability by text, and the compatibility with various personalized T2I models without specific tuning. 
To achieve these goals, \modelname\ builds upon a base T2I model with well-trained temporal alignment layers, allowing for the seamless transformation of any personalized T2I model into an image animation model. A key component of \modelname\ is the introduction of the condition module, which takes as inputs the condition frame and inter-frame affinity. This module leverages the affinity hint to transfer appearance information from the condition frame to individual frames in the latent space. This design mitigates the challenges of appearance-related frame alignment within \modelname\ and allows for a stronger focus on aligning with motion-related guidance. To address the lack of a benchmark for this field, we introduce \textbf{\benchname}, a comprehensive benchmark comprising diverse personalized T2I models, curated images, and motion-related prompts. 
We show extensive evaluations and applications on \benchname\ to verify the superiority of \modelname.
\end{abstract}

\vspace{-16mm}
\section{Introduction}
\label{sec:intro}
Recent democratization of text-to-image (T2I) generation has seen incredible progress by the growing Stable Diffusion community \cite{rombach2022high,automac111} and the explosion of personalized models \cite{ruiz2023dreambooth,hu2021lora}. Numerous AI artists and enthusiasts create, share, and use personalized models on model-sharing platforms such as Civitai \cite{civitai2022} and Hugging Face \cite{huggingface2022}, turning imagination into images at an affordable cost (e.g., a laptop with an RTX3080). However, there is still limited research exploring the realm of infusing these unique personalized images with dynamic motions \cite{wang2023videocomposer,guo2023animatediff,blattmann2023align}.

We are intrigued by the potential of animating personalized images, given their distinctive styles and impressive visual quality. Moreover, the incorporation of text prompts to guide the animation process can significantly streamline various creative endeavors within the vibrant personalized community, minimizing the required effort. While promising, this task presents two distinct challenges: \textbf{image alignment} with the detailed personalized image and achieving \textbf{motion controllability} through text prompt.
While recent advancements in training large-scale text-to-video (T2V) models have enabled users to generate videos based on text or images \cite{wang2023videocomposer}, these approaches struggle to preserve the unique styles and high-fidelity details of conditional images due to the absence of personalized domain knowledge.
To address this limitation, researchers have developed temporal alignment layers \cite{blattmann2023align,guo2023animatediff}. These plug-and-play modules effectively transform personalized T2I models into T2V models while maintaining their distinct styles \cite{blattmann2023align,guo2023animatediff}.
Despite numerous attempts to adapt these personalized T2V models for image animation \cite{zhang2023adding,ye2023ip-adapter}, 
the integrated temporal alignment layers mainly focus on aligning the appearance of individually generated frames in latent space to ensure smooth video outcomes. As a result, they find it hard to animate images according to motion-related text prompts.

To address the aforementioned limitations, we introduce \textbf{\modelname}, a \textbf{P}ersonalized \textbf{I}mage \textbf{A}nimator that excels in both image alignment and motion controllability.
First, we choose to build our framework upon a base T2I model (\ie, Stable Diffusion \cite{rombach2022high}) and incorporate well-established temporal alignment layers following previous works \cite{guo2023animatediff}. This approach allows \modelname\ to effectively leverage personalized domain knowledge by seamlessly replacing the base T2I model with any personalized T2I model during inference.
Second, to empower our framework with image animation ability, we introduce a trainable condition module into the input layer of the T2I model. This module plays a crucial role in generating each frame of the animation by taking the conditional image and the inter-frame affinity between the current frame and the conditional frame as inputs. Through this design, the condition module effectively borrows appearance features from the conditional frame, guided by the affinity hint, resulting in improved alignment with the condition image.
Finally, we fine-tune the condition module and the temporal alignment layers jointly while keeping the base T2I model fixed. 
Leveraging the enhanced appearance-related alignment facilitated by the condition module within the input layer, the temporal alignment layers can shift their focus towards motion-related alignment, leading to improved motion controllability.
Importantly, we demonstrate the effortless generalizability of \modelname's animation capability by replacing the T2I model with any other personalized model. This flexibility empowers users to animate their elaborated personal images using text prompts while preserving distinct features and high-fidelity details.

In summary, \modelname\ presents a powerful solution for personalized image animation, offering superior image alignment, motion controllability by text, and the flexibility to integrate various personalized models. This comprehensive approach ensures an engaging and customizable animation experience for users.
To address the lack of a benchmark in personalized image animation, we introduce a new benchmark called \textbf{\benchname}, which comprises various personalized T2Is, curated images, and tailored motion-related prompts.
To validate the effectiveness of \modelname, we conducted extensive quantitative and qualitative evaluations using \benchname. These evaluations provide a robust assessment of the capabilities and performance of our approach.

\figoverview 
\section{Related Work}
\label{sec:related}

\paragraph{Single Image Animation.}
Enhancing still pictures with motions has attracted decades of attention and efforts in the research field \cite{chuang2005animating,endo2019animating,bertiche2023blowing,xiao2023hairAni,holynski2021animating,shen2022learning,shen2023difftalk}. 
Chuang \etal present a pioneer work for animating a variety of photographs and paintings \cite{chuang2005animating}. Specifically, the still scene is segmented into a series of layers by a human user and each layer is rendered with a stochastic motion texture to form the final animated video. 
With the success of deep learning, more recent works get rid of the manual segmentation and can synthesize more natural motion textures \cite{endo2019animating,bertiche2023blowing,xiao2023hairAni,holynski2021animating,shen2022learning,shen2023difftalk}. 
Endo \etal propose learning and predicting the motion and appearance separately using two types of DNN models for animating landscape \cite{endo2019animating}. 
Holynski \etal learns an image-to-image translation network to encode motion priors of natural scenes with continuous fluid \cite{holynski2021animating}.
Bertiche \etal focus on learning garment motion dynamics for blowing the dressed humans under the wind \cite{bertiche2023blowing}.
Xiao \etal propose automatically segmenting hair wisps and animating hair wisps with a wisp-aware animation module with pleasing motions \cite{xiao2023hairAni}.
Some works focus on synthesizing natural lip motion for talking heads based on audio \cite{shen2023difftalk,shen2022learning}. 
Recent works also have been significantly prompted by the powerful diffusion models \cite{li2023generative,mahapatra2023synthesizing,rombach2022high}. Mahapatra \etal transfer the estimated optical flow from real images to artistic paintings by the concept of twin image synthesis with a pre-trained text-to-image diffusion model \cite{mahapatra2023synthesizing}. Li \etal leverages a latent diffusion model for modeling natural oscillating motion such as flowers and candles blowing in the wind \cite{li2023generative}. 
Despite the significant progress, these animation techniques can only synthesize specific types of content and motion such as time-lapse videos, and facial and body animation. In contrast, our model, \modelname, can animate the image without segmenting the region of interest and drive the image with any motion through a text prompt. 

\vspace{-3mm}
\paragraph{Text-to-video Synthesis.}
The synthesis of images and videos is a classic and important topic that has received significant research attention by decades \cite{zheng2020learning,ho2020ddpm,goodfellow2020generative,karras2019style}.
Recent large text-to-image models have shown remarkable progress by enabling diverse and high-fidelity image synthesis based on a text prompt written in natural language \cite{rombach2022high,saharia2022imagen,ramesh2021dalle1,ramesh2022dalle2,podell2023sdxl,zhuang2023task}. 
Most of text-to-video (T2V) generation models are built upon the well-trained text-to-image models and can synthesize smooth videos \cite{blattmann2023align,guo2023animatediff}. Some video latent diffusion models (VLDMs) have shown promising results in synthesizing natural videos \cite{wu2023tune, blattmann2023align, guo2023animatediff, esser2023structure,ho2022video,harvey2022flexible,luo2023videofusion,yang2023diffusion}. 
Despite numerous attempts that try to adapt these T2V models for image animation, it remains a challenge to align with the conditional image and control the image animation by text \cite{promptTravel,zhang2023adding}.  
Among them, we highlight AnimateDiff as a pioneer work that aims at training a plug-and-play motion module on large-scale video datasets \cite{guo2023animatediff}. After training, it can transfer the learned realistic motion priors to new domains by inserting the motion module into customized T2I models. 
Compared with AnimateDiff, our proposed PIA empowers any text-to-image models into an image animator, which can animate personalized images with better image fidelity and motion controllability.
\section{\modelname: Your Personalized Image Animator}

To animate an image $I$ generated by a personalized text-to-image model, our target is to generate a video clip $V=\{v^1,\dots, v^F\}$ with a length of $F$ frames, where the first frame $v^1$ should resemble condition image $I$, and the generated motion should follow the text prompt $c$. To achieve this goal, we propose \modelname, which aims at training plug-and-play models that is able to turn any text-to-image personalized models into an image animator.

As shown in \cref{fig:overview}, \modelname\ consists of a text-to-image model, well-trained temporal alignment layers, and a new condition module. First, the condition module takes as input the condition frame $I$ and the inter-frame affinity $s^{1:F}$. The condition module can transfer appearance information to each frame according to the affinity hints, leading to a better appearance alignment within the input layers. Second, the condition module and the temporal layers are trained jointly while maintaining the T2I model frozen. Such a design encourages the temporal alignment layers to shift focus toward motion-related alignment. Finally, \modelname\ can be combined with any personalized model by replacing the base T2I model. We introduce more in this section. 

\subsection{Preliminaries}\label{subsec:preliminaries}

\paragraph{Stable Diffusion.}
Stable Diffusion (SD) is one of the most popular large-scale open-source text-to-image models and has a well-developed community \cite{rombach2022high}. SD has four main components: a encoder $\mathcal{E} $, a decoder $\mathcal{D}$, a U-Net denoiser $\epsilon_{\theta}$, and a pretrained CLIP-based \cite{radford2021clip} text encoder $\mathcal{T}$.
In the training stage, the input image $x$ is first encoded to the latent code $z_0=\mathcal{E}(x)$. Then $z_0$ is perturbed via pre-defined diffusion process \cite{ho2020ddpm,nichol2021improved,dhariwal2021adm}:
\begin{equation}
    z_t = \sqrt{\bar{\alpha_t}} z_0 + \sqrt{1 - \bar{\alpha_t}} \epsilon_t, \epsilon_t \sim \mathcal{N}(0, I)
    \label{equ:diffusion}
\end{equation}
where timestep $t\sim \mathcal{U}[1, T]$, and $\bar{\alpha_t}$ is the noise strength at $t$.
Then the denoising U-Net $\epsilon$ is trained to predict the added noise with the following loss:
\begin{equation}
    \mathcal{L} = \mathbb{E}_{x,\epsilon\sim N(0, 1), t, c} \left[\Vert \epsilon - \epsilon_{\theta}(z_t, t, \mathcal{T}(c))  \Vert \right]
    \label{equ:loss_sd}
\end{equation}
where $c$ is the corresponding prompt for $x_0$.

\vspace{-3mm}
\paragraph{Temporal Alignment Layer.}
Recent T2V methods have demonstrated that incorporating attention layers \cite{vaswani2017attention} into SD in the temporal dimension enables the alignment of appearance and motion information across frames, thereby transforming the T2I model into a T2V model \cite{wu2023tune,guo2023animatediff,blattmann2023align}. We take a diffusion feature with a shape of $f\in\mathbb{R}^{F, C, H, W}$ as an example, where $F$ denotes the frame length, $C$ denotes the number of channels, and $H$ and $W$ denote the spatial dimensions. The temporal alignment layer begins by permuting the shape of the feature to $((H, W), F, C)$, and conducts the temporal attention as follows,
\begin{equation}
    f_{out} = \texttt{Softmax}(\frac{QK^{\intercal}}{\sqrt{c}}) \cdot V,
\end{equation}
where $Q=\mathcal{W}^Q z$, $K=\mathcal{W}^K z$, $V=\mathcal{W}^V z$ are projection operations. Through such a design, the temporal alignment layers are trained to align both appearance and motion information across all the frames simultaneously. Despite some promising results, a trade-off arises between appearance consistency and motion controllability in the generated videos.
In our work, \modelname\ introduces a condition module for appearance alignment and enables the temporal alignment layers to shift focus toward motion-related alignment.

\subsection{Plug-and-Play Modules for Animation}
\label{subsec:main_method}

\figConInExp
\modelname\ consists of a base text-to-image model, temporal alignment layers, and a condition module. It takes as input a condition frame and inter-frame affinities to generate videos. As illustrated in \cref{fig:explain_cond}-(a), in a typical T2V model, the initial latent for each frame is independently sampled from noise and synthesized individually, except for the temporal attention operations across all frames by the temporal alignment layers. As a result, temporal alignment layers play a key role as the only frame-aware modules, and it needs to devote much effort to simultaneously learning motion prior and aligning appearance consistency across frames. Such a design often leads to both inconsistent appearance identities for each frame and limited motion control by text prompt.

\vspace{-3mm}
\paragraph{Inter-frame Affinity.} 
To overcome the challenge of aligning both appearance and motion across all the frames simultaneously, we propose to explicitly encode the appearance into the latent from the condition frame $I$ with affinity hint, \ie $z^I = \mathcal{E}(I)$. 
During training, we calculate the affinity score based on the training data. Given a video clip $v^{1:F}$ and its first frame $v^1$ as the condition frame, we first calculate the L1 distance between each frame $v^i$ and the condition frame $v^1$ in HSV space, which is denoted as $d^i$. Next, we apply this operation to all frames of video clips in the dataset and find the maximum distance value $d_\mathrm{max}$. We normalize the distance $d^i$ to $[0, 1]$ via $d_\mathrm{max}$. Finally, the affinity score for each frame can be calculated by $s^i = 1 - d^i / d_\mathrm{max} \times (s_\mathrm{max} - s_\mathrm{min})$, where $s_\mathrm{max}$ and $s_\mathrm{min}$ are hyperparameters to scale affinity scores to specific ranges. To align the shape of $s^i$ with $z^I$, we expand it to a single-channel feature map with the same size as latent code (\ie $[1\times h \times w]$). After training, users can manually set the value of $s^i$ to animate images with varying motion scales, providing customization in the output.

\vspace{-3mm}
\paragraph{Condition Module.} 
To encode the additional inputs of the condition image and the inter-frame affinity based on the text-to-image U-Net, we introduce a lightweight learnable single-layer convolution as the condition module without compromising the original functionality. 
Given condition inputs $z^I \oplus s^i$, where $\oplus$ denotes concatenation operation, we encode it with the condition module $\mathcal{W}_{cond}$ and add the encoded feature to the output of the first convolution layer.

As shown in \cref{fig:explain_cond}-(b), the incorporation of the condition module leads to enhanced consistency of the appearance information in the output of the first convolutional layer compared to the original model. This makes subsequent temporal alignment layers shift focus toward aligning motion-related guidance. This operation is equivalent to concatenating $\mathcal{W}_{\mathrm{cond}}$ and $\mathcal{W}$, the weight of the original convolution block, at the first dimension. To insert the newly added weight without hurting the performance of the pre-trained model, we apply zero-initialization \cite{zhang2023adding} to $\mathcal{W}_{\mathrm{cond}}$.

\subsection{Training and inference}\label{subsec:train}
We inherit the pre-training of domain adapter and temporal alignment layers from AnimateDiff \cite{guo2023animatediff}. Specifically, the domain adapter pre-training is conducted on images from the video dataset to prevent the temporal alignment layers from learning the low visual quality of the video dataset. Then, the temporal alignment layers are trained on a video dataset to learn realistic motion prior. Finally, we insert the condition module into the first convolution layer in U-Net and initialize the weights as zero.
During training, given a sampled video clip $v^{1:F}$, we designate the first frame as the condition frame, \ie $I:=v^0$. We jointly trained the $\mathcal{W}_{\mathrm{cond}}$ and the temporal alignment layers while keeping the other parameters fixed as follows,
\begin{equation}
    \mathcal{L}=\mathbb{E}_{v^{1:F},\epsilon\sim \mathcal{N}(0, I),t, c}[\Vert \epsilon - \epsilon_\theta(z_t^{1:F}, z_0^0, s^{1:F},t,\mathcal{T}(c))||_2^2],
\end{equation}\label{equ:loss}
where $c$ is the text prompt, $z_0^0$ is the clean latent of the condition frame and $z_t^{1:F}$ is the perturbed latent.

During training, to retain the model's text-to-video capability, we randomly drop inputs with a 20\% probability, effectively transforming the training task into a T2V task. Specifically, we replace both $s^{1:F}$ and $z^{I}$ with zeros. When calculating the affinity score, we set $s_\mathrm{max}$ and $s_\mathrm{min}$ to be 1 and 0.2, respectively. Both single frame and video data training are conducted on WebVid dataset \cite{bain2021webvid}. The number of frames $F$ is fixed as 16 during training.
After training, given an image $I$ generated by a personalized model, we replace the domain adapter with the personalized model and fix the temporal alignment layers and condition module. The inter-frame affinity $s$ serves as a user input to control the magnitude of motion in generated videos.
\section{Experiments}
\label{sec:exp}

\subsection{AnimateBench}
\label{subsec:bench}

\paragraph{AnimateBench.} 
Existing benchmarks are restricted to specific domains like human faces, fluid elements, etc. To this end, we introduce \textbf{AnimateBench} for comparisons in the field of personalized image animation. AnimateBench contains images in different domains and multiple prompts to test the text-based image animation ability. Specifically, we evaluated 105 personalized cases, which contain different contents, styles, and concepts. These personalized cases are generated from seven different personalized text-to-image models, which are used to evaluate the domain generalization ability. We use five images generated by each model for a comprehensive comparison. In addition, we carefully elaborate three motion-related prompts for each image to evaluate the motion controllability by the text of different approaches. Specifically, the motion-related text prompts typically describe the following motions that the image probably happens within a single short shot.

\vspace{-3mm}
\paragraph{Evaluation Metrics.} 
Our quantitative comparison evaluates animation quality including \textbf{image alignment} and \textbf{text alignment}, following previous works \cite{wu2023tune,guo2023animatediff}. Specifically, the CLIP score is calculated by computing the cosine similarity between different embeddings, \eg, text and image. Therefore, we can use the CLIP score to evaluate both text and input image similarity with videos.
\begin{itemize}[nosep]
    \item \textbf{Image Alignment}  We compute the cosine similarity between input image embeddings and each video frame embeddings to evaluate the image-video alignment.
    \item \textbf{Text Alignment} We also compute the cosine similarity between text embeddings and frame embeddings to evaluate the text-video alignment.
\end{itemize}

\figanimatebench

\figvisual
\vspace{-3mm}
\paragraph{Baselines.}
We carefully choose the most recent and competitive approaches for personalized image animation with their brief introduction as follows,
\begin{itemize}[nosep]
    \item \textbf{VideoComposer} \cite{wang2023videocomposer} can generate controllable videos using different conditions such as text, image, motion vector, or a combination of them.
    \item \textbf{AnimateDiff} \cite{guo2023animatediff} is a milestone work that largely facilitates the development of personalized video generation by learning motion prior. We extend it for the application of image animation following previous best practices by using ControlNet \cite{zhang2023adding} and image prompt adapter \cite{ye2023ip-adapter}.
    \item \textbf{Pika Labs} \cite{pikalabs2023} and \textbf{Gen2} \cite{gen2} are two of the most popular commercial tools for video generation, which are widely recognized as state-of-the-art in image animation.
\end{itemize}

\subsection{Comparisons with State-of-the-Art}
\label{subsec:comp}
We compare our method with state-of-the-art for personalized image animation on \benchname. The performance of these models was evaluated in various terms of text alignment, and image alignment.

\vspace{-3mm}
\paragraph{Qualitative Comparison.}
We qualitatively compared PIA with the most recent animation methods, including AnimateDiff-ControlNet, Gen-2, Pika Labs, and VideoComposer. Visual results can be found in \cref{fig:visual}. It can be observed that the videos generated by PIA exhibit better responses to motion-related keywords in the prompts and better preservation of details from the input image content. VideoComposer produces video clips with poor text alignment and incoherent motion. Animatediff-ControlNet, on the other hand, generates videos with smooth motion but loses image details. Gen2 and Pika Labs are prone to generate videos with high fidelity but respond less to the prompt.

\vspace{-3mm}
\paragraph{Quantitative Evaluation.}
To test both the personalized cases in \benchname\ and open-domain cases in WebVid \cite{bain2021webvid} for a comprehensive comparison, we calculate the CLIP Score to compare text alignment and image alignment. We conclude the results in Table \ref{table:quant}. The results demonstrate that PIA achieves the best CLIP Scores, indicating strong text-video alignment and high image fidelity.

To conduct a user study, we generate videos with different methods and set up twenty questions in total. Note that, all the cases are selected randomly for fair testing. In each trial, we ask the users to choose the video that matches the text or image best. Finally, we conclude the preference rate to evaluate the performance of all the methods in \cref{table:quant}.

\tabquant

\subsection{Analysis}
\label{subsec:ablation}
In this section, we present some interesting applications and conduct ablative study.
Specifically, we first introduce three abilities of PIA, which are motion control by text prompt, motion magnitude controllability and style transfer. We show visual results for each application. Then we compare two different frame affinities to clarify the effectiveness. Details of different constructions of frame affinity will be explained in the following.

\vspace{-3mm}
\paragraph{Motion Control by Text Prompt.}
The condition module enables PIA to borrow appearance from the condition frame so that temporal alignment layers can focus more on motion generation. These designs enable PIA to perform better image and text alignment in generated videos. The strong prompt control ability of PIA can be found in \cref{fig:prompt}.
During the test, we observed that a simple prompt design produces great visual effects, such as fire and lightning. PIA can generate fancy results by adding new elements to the generated videos with smooth motion quality based on text prompts. Unlike Pika Labs or Gen-2 which are prone to generate small motion while ignoring text prompt, PIA generates video with more text alignment motion.

\figprompt
\figmotion
\vspace{-3mm}
\paragraph{Motion Magnitude Controllability.}
PIA can control the magnitude of motion by adjusting the frame affinity. 
Specifically, we first make statistics of the frame affinity across the whole video training dataset.
Then, according to the statistical results, we roughly divide the range of frame affinity into three groups and find three corresponding sets of inter-frame affinity values.
Finally, during inference, given the same condition image and text prompt, PIA can achieve videos with different motion magnitudes by using different sets of inter-frame affinity inputs.
As shown in \cref{fig:motion}, the frame affinity gradually decreases from top to bottom, indicating an increasing magnitude of motion. From the results, it can be observed that the dog can jump with different magnitudes, demonstrating the effectiveness of the frame affinity in controlling motion magnitude. 
In short, by using different frame affinity during the inference stage, users can flexibly control the motion magnitude of the generated video, leading to a more controllable animation.

\vspace{-3mm}
\paragraph{Style Transfer.}
So far, we have discussed using models and input images in the same domain to generate videos. However, we have found interesting facts that if the two have different domains, such as providing a real-style image and using an anime-style model to generate motion, PIA can achieve style transfer effects. As shown in \cref{fig:style}, frame affinity helps generate videos with smooth style transfer between image style and model style. This feature of PIA allows users to have a better personalized experience.
\figstyle

\vspace{-3mm}
\paragraph{Ablation Study on Frame Affinity.}
During training, we calculate the similarity between the condition frame and each frame as the affinity hints.
To verify the effectiveness of the inter-frame affinity design, we conduct an ablation study by comparing it with one-hot frame affinity. Specifically, one-hot frame affinity is constructed by setting 1 in the condition frame and setting 0 in the other frames. Such a design can indicate the position of the condition frame. 
We evaluate two methods on the AnimateBench. One-hot frame affinity obtains the image CLIP score of \textbf{210.9}, while the inter-frame affinity reaches \textbf{225.9}. Moreover, we show the visual result of video frames generated by these two designs in \cref{fig:affinity}. We can find that frame affinity constructed by similarity allows the model to align with the condition frame better. Note that in different constructions, we use the same pattern of affinity in training and inference. 
\figaffinity

\figlimit
\vspace{-3mm}
\paragraph{Limitation.}
First, we have observed that videos generated by PIA exhibit color discrepancy when applied to images with significantly different styles from the training data, as shown in \cref{fig:limit}. 
We hypothesize the color discrepancy is caused by training on the WebVid dataset \cite{bain2021webvid}. WebVid consists mainly of real-world recordings, characterized by frames with motion blur and compression artifacts \cite{bain2021webvid,guo2023animatediff}. Hence, the quality and diversity of these frames are lower than image datasets that include professional photography and artistic paintings \cite{blattmann2023align,guo2023animatediff}. Training with this noticeable quality domain gap can inevitably limit the animation of complex images with significant stylistic differences.

Secondly, in personalized image generation, users rely on specific trigger words in text prompts to evoke desired styles. However, during image animation, if these trigger words are absent from the animation prompts, significant color inconsistencies may arise.
We further speculate that training the model on a wider range of video data encompassing diverse styles and content can ease this issue. Additionally, ensuring the inclusion of complete trigger words during inference could help mitigate this phenomenon.

\section{Conclusions}
\label{sec:conclusion}
In this paper, we introduce \modelname\, a powerful solution for personalized image animation. Our method shows excellent image alignment and motion controllability, ensuring text-based customized animation for users. 
Furthermore, we construct an animation benchmark named \benchname\ to evaluate the animation performance of \modelname\ and other methods. Experimental results demonstrate that \modelname\ performs excellent on image animation tasks and shows various interesting extended applications.

\section{Acknowledgement}
This project is supported by the National Key R\&D Program of China (No. 2022ZD0161600)

{
    \small
    \bibliographystyle{ieeenat_fullname}
    \bibliography{main}
}
\newpage

\section{Implementation Details}
\label{sec:supp_imp}
\subsection{Training}
We train PIA on WebVid10M \cite{bain2021webvid} with only condition module and temporal alignment layers trainable.
Specifically, we use the motion module in AnimateDiff \cite{guo2023animatediff} as a pre-trained model for temporal alignment layers.
We compute the L1 distance between the condition frame and other frames in HSV space.
Subsequently, we utilize this distance to calculate the affinity score. 
Specifically, we consider the top 2.5th percentile of the samples as the minimum 
and the 97.5th as the maximum value to linearly scale the affinity score to $[0.2, 1]$.
In addition, we set a probability of $20\%$ to zero out the entire input of the condition module. 
This ensures that PIA retains text-to-video capabilities and promotes the training of the condition module.
We train condition module and temporal alignment layers on 16 NVIDIA A100s for 4.5k steps and use a learning rate of $1 \times 10^{-5}$.

\subsection{Inference}
During the inference stage, users can replace the base model with the personalized T2I model to realize the image animation.
Besides, we construct the inter-frame affinity 
according to the affinity score obtained from the training stage.
We design three affinity ranges for three different amplitude motions.
The maximum value of all three affinity ranges is 1, achieved at the conditional frame.
The minimum values are 0.2, 0.4, and 0.8, respectively, with corresponding decreases in motion magnitude.
We use classifier-free guidance during the DDIM process \cite{song2020ddim} and set the classifier-free guidance \cite{ho2022cfg} scale as 7.5. 
A 512 × 512 image can be animated in around 13.8 seconds (using 25 denoising steps with classifier guidance) on a single A100 GPU.

\section{AnimateBench}
\label{sec:supp_bench}
AnimateBench is a comprehensive benchmark, which consists of 105 image and prompt pairs.
To cover a wide variety of contents, styles, and concepts, we choose seven base models \cite{ruiz2023dreambooth} and LoRA 
 \cite{hu2021lora}.
An example case of AnimateBench is depicted in \cref{fig:benchcase}.
We have released AnimateBench in \href{https://huggingface.co/datasets/ymzhang319/AnimateBench}{https://huggingface.co/datasets/ymzhang319/AnimateBench}.

\figsuppbenchcase
\subsection{Images in AnimateBench}
We carefully choose seven of the most popular base models \cite{ruiz2023dreambooth} and LoRAs \cite{hu2021lora} in Cvitai \cite{civitai2022}. Each personalized model has very distinct styles and we use them to curate images with impressive high quality by tailored text prompts for image generation. Specifically, these images differ in styles, contents, and concepts and ensure that \benchname\ covers three categories: people, animals, and landscapes.

\figsuppluserstudy
\subsection{Prompts in AnimateBench}
For each generated image, we design three prompts describing different motions to test the text alignment ability of models.
Prompts are mainly composed of three parts: the \textbf{subject}, the \textbf{motion descriptor}, and the \textbf{trigger words}.
Subject and motion descriptors specify the content of motion in the generated videos.
The trigger word is a well-known technique that is able to activate the DreamBooth or LoRA to generate personalized effects \cite{civitai2022}. 
Only when these prompts are included during inference, DreamBooth or LoRA can achieve optimal performance.
Then we can get the complete prompt in AnimateBench.
For example, we use `1girl is smiling, white hair by atey ghailan, by greg rutkowski, by greg tocchini.' to generate a personalized image, and then we can get the complete prompt as \textit{`1girl is smiling, white hair by atey ghailan, by greg rutkowski, by greg tocchini'}.
In this case, \textit{`1gril'} represents the \textbf{subject}, \textit{smiling} represents the \textbf{motion descriptor}, and \textit{`white hair by atey ghailan, by greg rutkowski, by greg tocchini'} represents the \textbf{trigger word}.
We also distinguish motion between different types of subjects.
For example, the prompt of people or animals contains more descriptors such as smiling, crying, etc, 
while the prompt of landscapes or scenes contains more like raining, lightning, etc.

\section{Evaluation Details}
\label{sec:supp_eval}
\subsection{CLIP Score}
Following previous work \cite{guo2023animatediff, wang2023videocomposer, blattmann2023align}, we compute CLIP score to quantitatively evaluate the alignment in generated videos. 
In addition to calculating text alignment, 
we measure image alignment by computing the similarity between the embeddings of the generated video frames and the input images.
The two average CLIP scores are calculated on AnimateBench which contains 1680 frames.
We leverage the code provided by \cite{hessel2021clipscore} and use ViT-B/32 \cite{radford2021learning} model to extract the embedding of images and prompts.

\subsection{User Study}
For user study, we randomly select input image and prompt pairs in AnimateBench and then generate videos by using PIA, 
VideoComposer \cite{wang2023videocomposer} and AnimateDiff \cite{guo2023animatediff, promptTravel} with ControlNet \cite{zhang2023adding} and IP-Adapter \cite{ye2023ip-adapter}.
We ask the participants to choose from the three generated videos with the best image alignment or text alignment in each question. 
We show an illustration of the question cases in \cref{fig:suppluserstudy}.
There are 20 questions in total, and the order of options is shuffled.
A total of 20 participants were involved in the survey.
Following the previous work \cite{blattmann2023align}, we calculated the preference rate, and the results are shown in the main paper.

\section{Ablation}
\label{sec:supp_ab}
In this section, we introduce more ablation studies to verify the effectiveness of the inter-frame affinity and the fine-tuning of temporal alignment layers.

\subsection{Inter-frame Affinity}

To further verify the effectiveness of inter-frame affinity, we train a model without affinity hints for ablation study. 
We remove the affinity channel from the input of the condition module and the result is shown in \cref{fig:affinityablation}.
Compared to our method, videos generated by the model without inter-frame affinity are more incoherent and change suddenly.
\figaffinityablation

\subsection{Fine-tuning Temporal Alignment Layers}
In the training stage, we train both the condition module and temporal alignment layers.
We now show the result of only training the condition module with temporal alignment layers frozen in \cref{fig:onlyconditionmodule}. 
The result proves that the frozen temporary alignment layers failed to align the condition frame with other frames. 
\figonlyconditionmodule

\section{Visualization Results}
\label{sec:supp_vis}

\subsection{Visualization of Attention Map}
To demonstrate that the motion alignment of PIA is better than other methods, we visualize the average cross attention map of \textbf{motion descriptor} token.
We use prompt \textit{`the rabbit on is on fire'} as an example and visualize the cross attention map corresponding to token \textit{`fire'}, as shown in \cref{fig:featuremap}.
We can observe that in our method, the region attended by the \textit{`fire'} matches the region of flames.
In contrast, the motion descriptor token in the baseline method randomly attends to the entire context and cannot form a meaningful structure.
This phenomenon demonstrates that our method exhibits better motion alignment performance.

\figfeaturemap
\figcomposite
\figipadapter
\subsection{\modelname\ with complex prompts}
The temporal layers of PIA focus more on motion-related alignment which leading to improved motion controllability. Therefore, PIA is capable of responding to complex motion descriptions in prompts. We include composite animations in \cref{fig:composite}.
\subsection{\modelname\ for Open-Domain Images}
In this section, we further explore animating open-domain images with \modelname\ without using personalized T2I models. 
To further enhance the preservation of the information and details of the condition frame, we combine the Image Prompt Adapter (IP-Adapter) \cite{ye2023ip-adapter} with \modelname.
Specifically, we use a CLIP image encoder to extract features from the input images. 
Then, these image features are incorporated into each frame through the cross-attention mechanism in the UNet.
As shown in \cref{fig:ipadapter}, without using personalized models, our model successfully animates an open-domain image with realistic motion by text while preserving the identity of the given image.

\end{document}